\pdfoutput=1
\documentclass[11pt,letterpaper]{article}
\usepackage{emnlp2016}
\usepackage{times}
\usepackage{latexsym}

\usepackage{color,soul} 
\usepackage{dcolumn}
\newcolumntype{d}[1]{D{.}{.}{#1}}

\usepackage{graphicx}
\usepackage{algorithm}
\usepackage{algorithmic}
\usepackage{multirow}
\usepackage{rotating}
\usepackage{xcolor}

\usepackage{amsmath}
\usepackage{amsfonts}
\usepackage{xspace}
\usepackage{tikz}
\usetikzlibrary{calc}
\usetikzlibrary{decorations.pathmorphing}
\usepackage{booktabs}
\usepackage[warn]{textcomp}
\usepackage{booktabs}
\usepackage{paralist}
\usepackage[normalem]{ulem}
\usepackage{soul}

\newcommand\Mu{\mathbf{U}}
\newcommand\Mv{\mathbf{V}}
\newcommand\Mc{\mathbf{C}}

\newcommand\vu[1]{\mathbf{u}_{#1}}
\newcommand\vv[1]{\mathbf{v}_{#1}}
\newcommand\vh{\mathbf{h}}
\newcommand\vs{\mathbf{s}}
\newcommand\ocal{\mathcal{O}}

\emnlpfinalcopy

\def\emnlppaperid{***}

\title{Learning Crosslingual Word Embeddings without Bilingual Corpora}
\newcommand{\spadeaff}{\ensuremath{1}\xspace}
\newcommand{\clubaff}{\ensuremath{2}\xspace}

\author{Long Duong,$^{\spadeaff\clubaff}$ Hiroshi Kanayama,$^{\ensuremath{3}}$ Tengfei Ma,$^{\ensuremath{3}}$ Steven Bird$^{\spadeaff\ensuremath{4}}$ \and Trevor Cohn$^{\spadeaff}$\\ 
  $^{\spadeaff}$Department of Computing and Information Systems, University of Melbourne\\
  $^{\clubaff}$National ICT Australia, Victoria Research Laboratory\\
  $^{\ensuremath{3}}$IBM Research - Tokyo\\
  $^{\ensuremath{4}}$International Computer Science Institute, University of California Berkeley\\[-0.0ex]}

\date{}

\begin{document}

\maketitle

\begin{abstract}
Crosslingual word embeddings represent lexical items from different languages in the same vector space, enabling transfer of NLP tools. 
However, previous attempts had 
expensive resource requirements, difficulty incorporating monolingual data or were unable to handle polysemy.
We address these drawbacks in our method which takes advantage of a high coverage dictionary in an EM style training algorithm over monolingual corpora in two languages. 
Our model achieves state-of-the-art performance on bilingual lexicon induction task exceeding models using large bilingual corpora, and
competitive results on the monolingual word similarity and cross-lingual document classification task.  
\end{abstract}

\section{Introduction}
Monolingual word embeddings have had widespread success in many NLP tasks including sentiment analysis~\cite{socher-EtAl:2013:EMNLP}, dependency parsing~\cite{dyer-EtAl:2015:ACL-IJCNLP}, machine translation~\cite{DBLP:journals/corr/BahdanauCB14}.  
Crosslingual word embeddings are a natural extension facilitating various crosslingual tasks, e.g. through transfer learning. 
A model built in a source resource-rich language can then applied to the target resource poor languages~\cite{YarowskyAndNgai,Das:2011,Tackstrom:2012:CWC,duong-EtAl:2015:ACL-IJCNLP}. 
A key barrier for crosslingual transfer is lexical matching between the source and the target language. 
Crosslingual word embeddings are a natural remedy where both source and target language lexicon are presented as dense vectors in the same vector space~\cite{klementiev-titov-bhattarai:2012}. 

Most previous work has focused on down-stream crosslingual applications such as document classification and dependency parsing. We argue that good crosslingual embeddings should preserve both monolingual and crosslingual quality which we will use as the main evaluation criterion through monolingual word similarity and bilingual lexicon induction tasks.  Moreover, many prior work~\cite{Chandar-nips-14,kovcisky-hermann-blunsom:2014:P14-2} used bilingual or comparable corpus which is also expensive for many low-resource languages. 
\newcite{sogaard-EtAl:2015:ACL-IJCNLP} impose a less onerous data condition in the form of linked Wikipedia entries across several languages, however this approach tends to underperform other methods.
To capture the monolingual distributional properties of words it is crucial to train on large monolingual corpora~\cite{Luong-etal:naacl15:bivec}. 
However, many previous approaches are not capable of scaling up either because of the complicated objective functions or the nature of the algorithm. 
Other methods use a dictionary as the bridge between languages \cite{DBLP:journals/corr/MikolovLS13,W14-1613}, however they do not adequately handle translation ambiguity.

Our model uses a bilingual dictionary from Panlex~\cite{Kamholz14} as the source of bilingual signal. 
Panlex covers more than a thousand languages and therefore our approach applies to many languages, including low-resource languages. 
Our method selects the translation based on the context in an Expectation-Maximization style training algorithm which explicitly handles polysemy through incorporating multiple dictionary translations (word sense and translation are closely linked~\cite{Resnik:1999:DSD:973980.973981}).
In addition to the dictionary, our method only requires monolingual data, as an extension of the continuous bag-of-word (CBOW) model~\cite{mikolov-yih-zweig:2013:NAACL-HLT}. 
We experiment with several variations of our model, whereby we predict only the translation or both word and its translation and consider different ways of using the different learned center-word versus context embeddings in application tasks.
We also propose a regularisation method to combine the two embedding matrices during training. 
Together, these modifications substantially improve the performance across several tasks. 
Our final model achieves state-of-the-art performance on bilingual lexicon induction task, large improvement over word similarity task compared with previous published crosslingual word embeddings, and competitive result on cross-lingual document classification task. 
Notably, our embedding combining techniques are general, yielding improvements also for monolingual word embedding. Our contributions are:
\begin{itemize}
\itemsep0em
\item Propose a new crosslingual training method for learning vector embeddings, based only on monolingual corpora and a bilingual dictionary. 
\item Evaluate several methods for combining embeddings which help in both crosslingual and monolingual evaluations.
\item Achieve uniformly excellent results which are competitive in monolingual, bilingual and crosslingual transfer settings. 
\end{itemize}

\section{Related work}
\label{sec:rel_work}
There is a wealth of prior work on crosslingual word embeddings, which all exploit some kind of bilingual resource.
This is often in the form of a parallel bilingual text, using word alignments as a bridge between tokens in the source and target languages, such that translations are assigned similar embedding vectors~\cite{Luong-etal:naacl15:bivec,klementiev-titov-bhattarai:2012}. 
These approaches are affected by errors from automatic word alignments, motivating other approaches which operate at the sentence level~\cite{Chandar-nips-14,DBLP:journals/corr/HermannB14,icml2015_gouws15} through learning compositional vector representations of sentences, in order that sentences and their translations representations closely match.
The word embeddings learned this way capture translational equivalence, despite not using explicit word alignments.
Nevertheless, these approaches demand large parallel corpora, which are not available for many language pairs.

\newcite{vulic-moens:2015:ACL-IJCNLP} use bilingual comparable text, sourced from Wikipedia. 
Their approach creates a psuedo-document by forming a bag-of-words from the lemmatized nouns in each comparable document concatenated over both languages.
These pseudo-documents are then used for learning vector representations using \texttt{Word2Vec}.
Their system, despite its simplicity, performed surprisingly well on a bilingual lexicon induction task (we compare our method with theirs on this task.)
Their approach is compelling due to its lesser resource requirements, although comparable bilingual data is scarce for many languages. 
Related,~\newcite{sogaard-EtAl:2015:ACL-IJCNLP} exploit the comparable part of Wikipedia. They represent word using Wikipedia entries which are shared for many languages. 

A bilingual dictionary is an alternative source of bilingual information.
\newcite{gouws-sogaard:2015:NAACL-HLT} randomly replace the text in a monolingual corpus with a random translation, using this corpus for learning word embeddings. 
Their approach doesn't handle polysemy, as very few of the translations for each word will be valid in context. 
For this reason a high coverage or noisy dictionary with many translations might lead to poor outcomes.
\newcite{DBLP:journals/corr/MikolovLS13},~\newcite{W14-1613} and~\newcite{faruqui-dyer:2014:EACL} filter a bilingual dictionary for one-to-one translations, thus side-stepping the problem, however discarding much of the information in the dictionary. 
Our approach also uses a dictionary, however we use all the translations and explicitly disambiguate translations during training. 

Another distinguishing feature on the above-cited research is the method for training embeddings.
\newcite{DBLP:journals/corr/MikolovLS13} and~\newcite{faruqui-dyer:2014:EACL} use a cascade style of training where the word embeddings in both source and target language are trained separately and then combined later using the dictionary.
Most of the other works train multlingual models jointly, which appears to have better performance over cascade training~\cite{icml2015_gouws15}. 
For this reason we also use a form of joint training in our work.

\section{Word2Vec}

Our model is an extension of the contextual bag of words (CBOW) model of~\newcite{mikolov-yih-zweig:2013:NAACL-HLT},
a method for learning vector representations of words based on their
distributional contexts.
Specifically, their model describes the probability of a token $w_i$ at
position $i$ using logistic regression with a factored
parameterisation,
\begin{equation}
p(w_i|w_{i \pm k \backslash i}) = \frac{\exp (\vu{w_i}^\top \vh{}_i)}{\sum_{w \in W} \exp (\vu{w}^\top \vh{}_i) }  \, ,
\label{equ:w2vpre}
\end{equation}
where $\vh{}_i = \frac{1}{2k} \sum_{j=-k; j \neq 0}^k \vv{w_{i+j}} $ is a vector encoding  the context over a window of size $k$ centred around position $i$,
$W$ is the vocabulary and
the parameters $\Mv$~and~$\Mu \in \mathbb{R}^{|W| \times d}$  are matrices referred to as the
context and word embeddings.
The model is trained to maximise the log-pseudo likelihood of a training corpus, however due to the high complexity of computing the denominator of \eqref{equ:w2vpre}, \newcite{mikolov-yih-zweig:2013:NAACL-HLT} propose  negative sampling as an approximation, by instead learning to differentiate data from noise (negative examples).
This gives rise to the following optimisation objective
\begin{equation}
\!\! \sum_{i \in D} \! \bigg( \log\sigma (\vu{w_i}^\top \vh{}_i) \!+\! \sum_{j=1}^p \mathbb{E}_{w_j \sim P_n(w)} \log\sigma (-\vu{w_j}^\top \vh{}_i) \bigg)
\label{equ:w2v_neg} \, ,
\end{equation}
where $D$ is the training data and $p$ is the number of negative examples randomly drawn from a noise distribution $P_n(w)$. 

\section{Our Model}

Our approach extends CBOW to model bilingual text, using two monolingual corpora and a bilingual dictionary. 
We believe this data condition to be less stringent than requiring parallel or comparable texts as the source of the bilingual signal.
It is common for field linguists to construct a bilingual dictionary when studying a new language, as one of the first steps in the language documentation process. 
Translation dictionaries are a rich information source, capturing much of the lexical ambiguity in a language through translation. 
For example, the word \textit{bank} in English might mean the \textit{river bank} or \textit{financial bank} which corresponds to two different translations \textit{sponda} and \textit{banca} in Italian.
If we are able to learn to select good translations, then this implicitly resolves much of the semantic ambiguity in the language, and accordingly we seek to use this idea to learn better semantic vector representations of words. 

\subsection{Dictionary replacement}
\label{sec:dict_rep}

To learn bilingual relations, we use the context in one language to predict the translation of the centre word in another language. 
This is motivated by the fact that the context is an excellent means of disambiguating the translation for a word. 
Our method is closely related to~\newcite{gouws-sogaard:2015:NAACL-HLT}, however we only replace the middle word $w_i$ with a translation $\bar{w_i}$ while keeping the context fixed.
We replace each centre word with a translation on the fly during training, predicting instead
$p(\bar{w}_i|w_{i \pm k \backslash i})$ but using the same formulation as (\ref{equ:w2vpre}) albeit with an augmented $\Mu$ matrix to cover word types in both languages.

\newcommand\uv{\theta}

\begin{algorithm}[t]
\begin{algorithmic} [1]
\STATE randomly initialize $\Mv$, $\Mu$  
\FOR{ $i< \texttt{Iter}$}
	\FOR {$i \in D_e \cup D_f $}
		\STATE $\vs \gets \vv{w_i} + \vh{}_i$ 
		\STATE $\bar{w_i} = \operatorname{argmax}_{w \in \text{dict}(w_i)} \cos(\vs,\vv{w}) $
		\STATE $\uv \gets \uv + \eta \frac{\partial \ocal(\bar{w_i}, w_i, \vh{}_i)}{\partial \uv}$
\COMMENT{see (\ref{equ:w2v_neg_joint}) or (\ref{equ:reg_combine})}
	\ENDFOR
\ENDFOR
\end{algorithmic}
\caption{EM algorithm for selecting translation during training,
where \mbox{$\theta = (\Mu, \Mv)$}  are the model parameters 
and $\eta$ is the learning rate.}
\label{alg:em_translation}
\end{algorithm}

The translation $\bar{w_i}$ is selected from the possible translations of $w_i$ listed in the dictionary.
The problem of selecting the correct translation from the many options is reminiscent of the problem faced in expectation maximisation (EM), in that cross-lingual word embeddings will  allow for accurate translation, however to learn these embeddings we need to know the translations.
We propose an EM-inspired algorithm, as shown in Algorithm~\ref{alg:em_translation}, which 
operates over both monolingual corpora, $D_e$ and $D_f$. 
The vector $\vs$ is the semantic representation combining both the centre word, $w_i$, and the context,\footnote{Using both embeddings gives a small improvement compared to just using context vector $\vh$ alone.} 
which is used to choose the best translation into the other language from the bilingual dictionary $dict(w_i)$.\footnote{We also experimented with using expectations over translations, as per standard EM, with slight degredation in results.}
After selecting the translation, we use $\bar{w_i}$ together with the context vector $\vh$ to make a stochastic gradient update of the CBOW log-likelihood.

\subsection{Joint Training}
\label{sec:join_training}
Words and their translations should appear in very similar contexts. 
One way to enforce this is to jointly learn to predict both the word and its translation from its monolingual context.  
This gives rise to the following joint objective function, 
\begin{multline}
\!\!\!\!\ocal = \! \sum_{i\in D_e \cup D_f} \bigg( \alpha  \log\sigma (\vu{w_i}^\top \vh{}_i) + (1 - \alpha)   \log\sigma (\vu{\bar{w_i}}^\top \vh{}_i) \\
+ \sum_{j=1}^p \mathbb{E}_{w_j \sim P_n(w)} \log\sigma (-\vu{w_j}^\top \vh{}_i) \bigg)\, ,
\label{equ:w2v_neg_joint}
\end{multline}
where $\alpha$ controls the contribution of the two terms. For our experiments, we set  $\alpha = 0.5$. The negative examples are drawn from combined vocabulary unigram distribution calculated from combined data $D_e \cup D_f$. 

\subsection{Combining Embeddings}
\label{sec:combine_emb}
Many vector learning methods learn two embedding spaces $\Mv$ and $\Mu$. Usually only $\Mv$ is used in application. The use of $\Mu$, on the other hand, is under-studied~\cite{DBLP:conf/nips/LevyG14} with the exception of~\newcite{pennington2014glove} who use a linear combination $\Mu + \Mv$, with minor improvement over $\Mv$ alone. 
 
\begin{figure}
 \centering
 \includegraphics[scale=0.4]{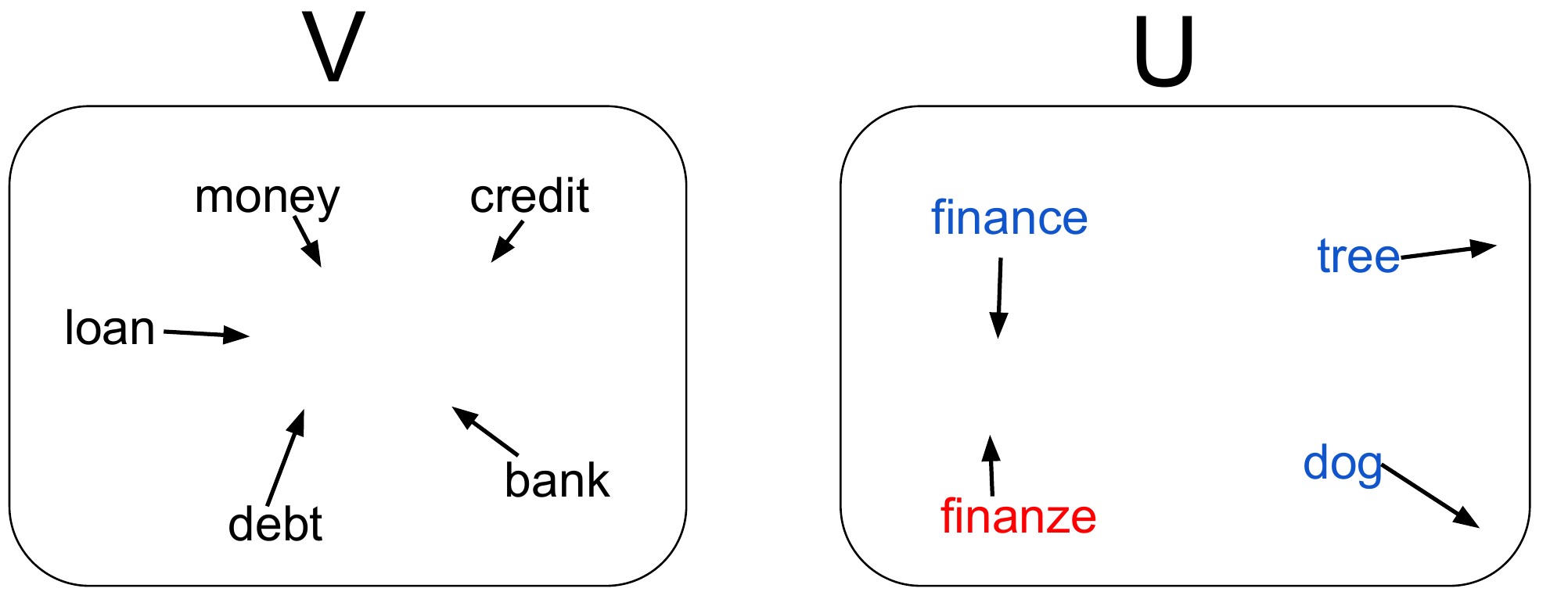}
 \caption{Example of $\Mv$ and $\Mu$ space during training.}
 \label{fig:combine_spaces}
 \end{figure}

We argue that with our model 
$\Mv$ is better at capturing the monolingual regularities and $\Mu$ is better at capturing bilingual signal. 
The intuition for this is as follows. Assuming that we are predicting the word \textit{finance} and its Italian translation \textit{finanze} from the context  (\textit{money, loan, bank, debt, credit}) as shown in figure~\ref{fig:combine_spaces}. In $\Mv$ only the context word representations are updated and in $\Mu$ only the representations of \textit{finance, finanze} and negative samples such as \textit{tree} and \textit{dog} are updated. CBOW learns good embeddings because each time it updates the parameters, the words in the contexts are pushed closer to each other in the $\Mv$ space. Similarly, the target word $w_i$ and the translation $\bar{w_i}$ are also pushed closer in the $\Mu$ space. This is directly related to poitwise mutual information values of each pair of word and context explained in~\newcite{DBLP:conf/nips/LevyG14}. 
Thus, $\Mu$ is bound to better at bilingual lexicon induction task and $\Mv$ is better at monolingual word similarity task. 

The simple question is, how to combine both $\Mv$ and $\Mu$ to produce a better representation. 
We experiment with several ways to combine $\Mv$ and $\Mu$. First, we can follow~\newcite{pennington2014glove} to \textit{interpolate} $\Mv$ and $\Mu$ in the post-processing step. i.e. 
\begin{equation}
\gamma   \Mv + (1-\gamma)  \Mu
\label{equ:interpolate_combine}
\end{equation} 
where $\gamma$ controls the contribution of each embedding space.
Second, we can also \textit{concatenate} $\Mv$ and $\Mu$ instead of interpolation such that $\Mc = [\Mv : \Mu]$ where $\Mc \in \mathbb{R}^{|W| \times 2d}$ and $W$ is the combined vocabulary from $D_e \cup D_f$. 
 
Moreover, we can also fuse $\Mv$ and $\Mu$ during training. 
For each word considered in equation~\ref{equ:w2v_neg_joint} in $\Mu$ space including $W_r = \{w_i,\bar{w_i}, w_j\}$ with $1\leq j \leq p$, we encourage the model to learn similar representation in both $\Mv$ and $\Mu$ by adding  
a \textit{regularization} term to the objective function in equation~(\ref{equ:w2v_neg_joint}) during training.
\begin{equation}
\ocal = \ocal +  \delta  \sum_{x \in W_r}\| \vu{x} - \vv{x}  \|^2_2
\label{equ:reg_combine}
\end{equation}
where $\delta$ controls to what degree we should bind two spaces together.  
\section{Experiment Setup}
We want to test the cross-lingual property, monolingual property and the down-stream usefulness of our crosslingual word embeddings (CLWE). For the crosslingual property we adopt the bilingual lexicon induction task from~\newcite{vulic-moens:2015:ACL-IJCNLP}. For the monolingual property we adopt the word similarity task on common datasets such as WordSim353 and Rareword. To demonstrate the usefulness of our CLWE, we also evaluate on the conventional crosslingual document classification task. 
  
\subsection{Monolingual Data}
The monolingual data is mainly from the pre-processed Wikipedia dump from~\newcite{polyglot:2013:ACL-CoNLL}. The data is already cleaned and tokenized. We additionally low-cased all words. Normally, the monolingual word embeddings are trained on billions of words. However, getting that much of monolingual data for a low-resource language is also challenging. That is why we only select the top 5 million sentences (around 100 million words) for each language. 

\subsection{Dictionary}
The bilingual dictionary is the only source of bilingual correspondence in our technique. We want a dictionary that covers many languages so that our approach can be applied widely to many low-resource languages. 
We use Panlex, a dictionary which currently covers around 1300 language varieties with about 12 million expressions. 
The translations in PanLex come from various sources such as glossaries, dictionaries, automatic inference from other languages, etc. 
Accordingly, Panlex has high language coverage but often noisy translations.
\footnote{We also experimented with a growing crow-sourced dictionary from Wiktionary. 
Our initial observation is that the translation quality is better but lower-coverage. 
For example, for Engish - Italian dictionary, Panlex and Wiktionary has the coverage of 42.1\% and 16.8\% respectively for the top 100k most frequent English words from Wikipedia. The average number of translations are 5.2 and 1.9 respectively. We observed similar trend using Panlex and Wiktionary dictionary in our model. However, using Panlex results in much better performance. We can run the model on the combined dictionary from both Panlex and Wiktionary but we leave it for future work.} 

\section{Bilingual Lexicon Induction}
Given a word in a source language, the bilingual lexicon induction (BLI) task is to predict its translation in the target language.~\newcite{vulic-moens:2015:ACL-IJCNLP} proposed this task to test crosslingual word embeddings. The difficulty of this is that it is evaluated using recall at one where each term has only a single gold translation. The model must be very discriminative in order to score well. 

We build the CLWE for 3 language pairs: it - en, es - en and nl - en, 
using similar parameters setting with~\newcite{vulic-moens:2015:ACL-IJCNLP}.\footnote{Default learning rate of 0.025, negative sampling with 25 samples, subsampling rate of value $1e^{-4}$, embedding dimension $d = 200$, window size $cs = 48$ and run for 15 epochs.} The remaining tunable parameters in our system are $\delta$ from Equation~(\ref{equ:reg_combine}), and the choice of algorithm for combining embeddings (see ~\S\ref{sec:parameter_tuning}).
\paragraph{Qualitative evaluation}

\begin{table}
\centering
\resizebox{\columnwidth}{!}{
\begin{tabular}{ll|ll}
\toprule
\multicolumn{2}{c}{es (gravedad) - en } & \multicolumn{2}{c}{it (tassazione) - en } \\
es              & en                 & it & en \\ 
\midrule
gravitacional   & \textbf{gravity}      & tasse              & \textbf{taxation}     \\
gravitatoria    & gravitational    & fiscale            & taxes   \\
aceleración     & acceleration    & tassa              & tax  \\
gravitación     & non-gravitational   & imposte            & levied   \\
inercia         & inertia            & imposta            & fiscal\\
gravity         & centrifugal       & fiscali            & low-tax  \\
msugra          & free-falling   & l'imposta          & revenue  \\
centr{\'i}fuga      & gravitational     & tonnage            & levy  \\
curvatura       & free-fall          & tax                & annates   \\
masa            & newton        & accise             & evasion \\
\bottomrule
\end{tabular}
} 
\caption{Top 10 closest words in both source and target language corresponding to Spanish word \textit{gravedad} and Italian word \textit{tassazione}. The correct translation in English is bold. }
\label{tab:top10closest}
\end{table}

We jointly train the model to predict both $w_i$ and the translation $\bar{w_i}$, combine $\Mv$ and $\Mu$ during training with regularization sensitivity $\delta = 0.01$ and use $\Mu$ as the output for each language pair. 
Table~\ref{tab:top10closest} shows the top 10 closest words in both source and target languages according to cosine similarity. Note that the model correctly identifies the translation in English, and the top 10 words in both source and target languages are highly related. This qualitative evaluation initially demonstrates the ability of our CLWE to capture both the bilingual and monolingual relationship. 

\paragraph{Quantitative evaluation}
\begin{table*}
\centering
\begin{tabular}{lllllllll}
\toprule
Model                             & \multicolumn{2}{c}{es - en} & \multicolumn{2}{c}{it - en} & \multicolumn{2}{c}{nl - en} & \multicolumn{2}{c}{Average} \\
                                  & $rec_1$      & $rec_5$        &  $rec_1$      & $rec_5$        & $rec_1$      & $rec_5$  & $rec_1$      & $rec_5$         \\
\midrule
\newcite{gouws-sogaard:2015:NAACL-HLT} + Panlex & 37.6         & 63.6         & 26.6         & 56.3         & 49.8         & 76.0       &  38.0 & 65.3 \\
\newcite{gouws-sogaard:2015:NAACL-HLT}  + Wikt   & 61.6         & 78.9         & 62.6         & 81.1         & 65.6         & 79.7       & 63.3 & 79.9  \\
BilBOWA: \newcite{icml2015_gouws15}      & 51.6         & -            & 55.7         & -            & 57.5         & -           & 54.9 & - \\
\newcite{vulic-moens:2015:ACL-IJCNLP}             & 68.9         & -            & 68.3         & -            & 39.2         & -    & 58.8 & -        \\
\midrule
Our model (random selection)      & 41.1         & 62.0         & 57.4         & 75.4         & 34.3         & 55.5       & 44.3 & 64.3 \\
Our model (EM selection)          & 67.3         & 79.5         & 66.8         & 82.3         & 64.7         & 82.4         & 66.3 & 81.4\\
+ Joint model                     & 68.0         & 80.5         & 70.5         & 83.3         & 68.8         & 84.0         &  69.1 & 82.6\\
+ combine embeddings ($\delta = 0.01$) & 71.6         & 84.4         & 78.7         & 89.5         & 76.9         & 90.1   & 75.7 & 88.0      \\
+ lemmatization                       & \textbf{71.8}         & \textbf{85.0}         & \textbf{79.6}         & \textbf{90.4}         & \textbf{77.1}         & \textbf{90.6}  & \textbf{76.2} & \textbf{88.7}       \\
\bottomrule
\end{tabular}
\caption{Bilingual Lexicon Induction performance from Spanish, Italian and Dutch to English. \protect\newcite{gouws-sogaard:2015:NAACL-HLT} + Panlex/Wikt is our reimplementation using Panlex/Wiktionary dictionary. 
All our models use Panlex as the dictionary. We reported the recall at 1 and 5. The best performance is bold. }
\label{tab:bli_result}
\end{table*}
Table~\ref{tab:bli_result} shows our results compared with prior work. We reimplement~\newcite{gouws-sogaard:2015:NAACL-HLT} using Panlex  and Wiktionary dictionaries. The result with Panlex is substantially worse than with Wiktionary. This confirms our hypothesis in~\S\ref{sec:rel_work}. That is the context might be very biased if we just randomly replace the training data with the translation especially with noisy dictionary such as Panlex.

Our model when randomly picking the translation is similar to~\newcite{gouws-sogaard:2015:NAACL-HLT}, using the Panlex dictionary. 
The biggest difference is that they replace the training data (both context and middle word) while we fix the context and only replace the middle word. For a high coverage yet noisy dictionary such as Panlex, our approach gives better average score. 
Our non-joint model with EM to select the translation\footnote{Optimizing equation~(\ref{equ:w2v_neg_joint}) with $\alpha = 0$.}, out-performs just randomly select the translation by a significant margin.

Our joint model, as described in equation~(\ref{equ:w2v_neg_joint}) which  predicts both target word and the translation, further improves the performance, especially for Dutch. 
We use equation~(\ref{equ:reg_combine}) to combine both context embeddings $\Mv$ and word embeddings $\Mu$ for all three language pairs. This modification during training substantially improves the performance.
More importantly, all our improvements are consistent for all three language pairs and both evaluation metrics, showing the robustness of our models. 

Our combined model out-performed previous approaches by a large margin.~\newcite{vulic-moens:2015:ACL-IJCNLP} used bilingual comparable data, but this might be hard to obtain for some language pairs. Their performance on Dutch is poor because their comparable data between English and Dutch is small. Besides, they also use POS tagger and lemmatizer to filter only \textit{Noun} and reduce the morphology complexity during training. These tools might not be available for many languages. For a fairer comparison to their work, we also use the same Treetagger~\cite{Schmid95improvementsin} to lemmatize the output of our combined model before evaluation. Table~\ref{tab:bli_result} (+lemmatization) shows some improvements but minor. It demonstrates that our model is already good at disambiguating morphology. For example, the top 2 translations for Spanish word \textit{lenguas} in English are \textit{languages} and \textit{language} which correctly prefer the plural translation. 

\section{Monolingual Word Similarity}
\label{sec:mono}
Now we consider the efficacy of our CLWE on monolingual word similarity. 
Our experiment setting is similar with~\newcite{Luong-etal:naacl15:bivec}. We evaluated on English monolingual similarity on WordSim353 (WS-EN), RareWord (RW-En) and German version of WordSim353 (WS-De)~\cite{ws353,LuongSM13,Luong-etal:naacl15:bivec}. 
Each of those datasets contain many tuples $(w_1,w_2,\texttt{score})$ where \texttt{score} is given by annotators showing the semantic similarity between $w_1$ and $w_2$. The system must give the score correlated with human judgment. 

\begin{table}
\centering
\tabcolsep 4pt
\resizebox{\columnwidth}{!}{
\begin{tabular}{lllll}
\toprule
& Model                      & WS-de & WS-en & RW-en \\
\midrule
\multirow{5}{*}{\rotatebox{90}{Baselines}}& \newcite{klementiev-titov-bhattarai:2012}       & 23.8     & 13.2     & 7.3         \\
& \newcite{Chandar-nips-14}        & 34.6     & 39.8     & 20.5        \\
& \newcite{DBLP:journals/corr/HermannB14} & 28.3     & 19.8     & 13.6        \\
& \newcite{Luong-etal:naacl15:bivec}         & 47.4     & 49.3     & 25.3        \\
& \newcite{gouws-sogaard:2015:NAACL-HLT}         & 67.4   & 71.8 & 31.0 \\
\midrule 
\multirow{3}{*}{\rotatebox{90}{Mono}} &  CBOW              & 62.2  & 70.3     & 42.7        \\
& + combine & 65.8 &  74.1 & 43.1 \\ 
& SOTA                       & -        & 81.0     & 48.3       \\
\midrule
\multirow{2}{*}{\rotatebox{90}{Ours}} & Our joint-model             & 59.3 & 68.6 & 38.1 \\ 
& + combine                 & \textbf{70.6}     & \textbf{75.7 }    & \textbf{44.6}        \\
\bottomrule
\end{tabular}
}
\caption{Spearman's rank correlation for monolingual similarity measurement on 3 datasets WS-de (353 pairs), WS-en (353 pairs) and RW-en (2034 pairs). We compare against 5 baseline crosslingual word embeddings. 
The best CLWE performance is bold. For reference, we add the monolingual CBOW with/without embeddings combination and monolingual SOTA result for each datasets:~\protect\newcite{Yih:2012:MWR:2382029.2382130} and~\protect\newcite{DBLP:journals/corr/ShazeerDEW16} for WS-en and RW-en}
\label{tab:mono_sim}
\end{table}

We train the model as described in equation~(\ref{equ:reg_combine}), which is exactly the same model as \textit{combine embeddings} in Table~\ref{tab:bli_result}.
Since the evaluation involves German and English word similarity, we train the CLWE for English - German pair. 
Table~\ref{tab:mono_sim} shows the performance of our combined model compared with several baselines. 
Our combined model out-performed both~\newcite{Luong-etal:naacl15:bivec} and~\newcite{gouws-sogaard:2015:NAACL-HLT}\footnote{use Panlex dictionary} which represent the best published crosslingual embeddings trained on bitext and monolingual data respectively.

We also train the monolingual CBOW model with the same parameter settings on the monolingual data for each language. 
Surprisingly, our combined model performs better than the monolingual CBOW baseline which makes our result closer to the monolingual state-of-the-art on each different dataset.
However, the best monolingual methods use massive monolingual data~\cite{DBLP:journals/corr/ShazeerDEW16}, WordNet and output of commercial search engines~\cite{Yih:2012:MWR:2382029.2382130}. 

Next we explain the gain of our combined model compared with the monolingual CBOW model. First, we compare the combined model with the joint-model w.r.t. monolingual CBOW model (Table~\ref{tab:mono_sim}). It shows that the improvement seems mostly come from combining $\Mv$ and $\Mu$. If we apply the combining algorithm to the monolingual CBOW model (CBOW + combine), we also observe an improvement. Clearly most of the improvement is from combining $\Mv$ and $\Mu$, however our $\Mv$ and $\Mu$ are much more complementary. The other improvements can be explained  by the observation that a dictionary can improve monolingual accuracy through linking synonyms~\cite{faruqui-dyer:2014:EACL}. For example, since \textit{plane}, \textit{airplane} and \textit{aircraft} have the same Italian translation \textit{aereo}, the model will encourage those words to be closer in the embedding space.
 
\section{Model selection}
\label{sec:parameter_tuning}
Combining context embeddings and word embeddings results in an improvement in both monolingual similarity and bilingual lexicon induction.  In \S\ref{sec:combine_emb}, we introduce several combination methods including post-processing (interpolation and concatenation) and during training (regularization). In this section, we justify our parameter and model choices. 
 \begin{figure}
 \centering
 \includegraphics[scale=0.5]{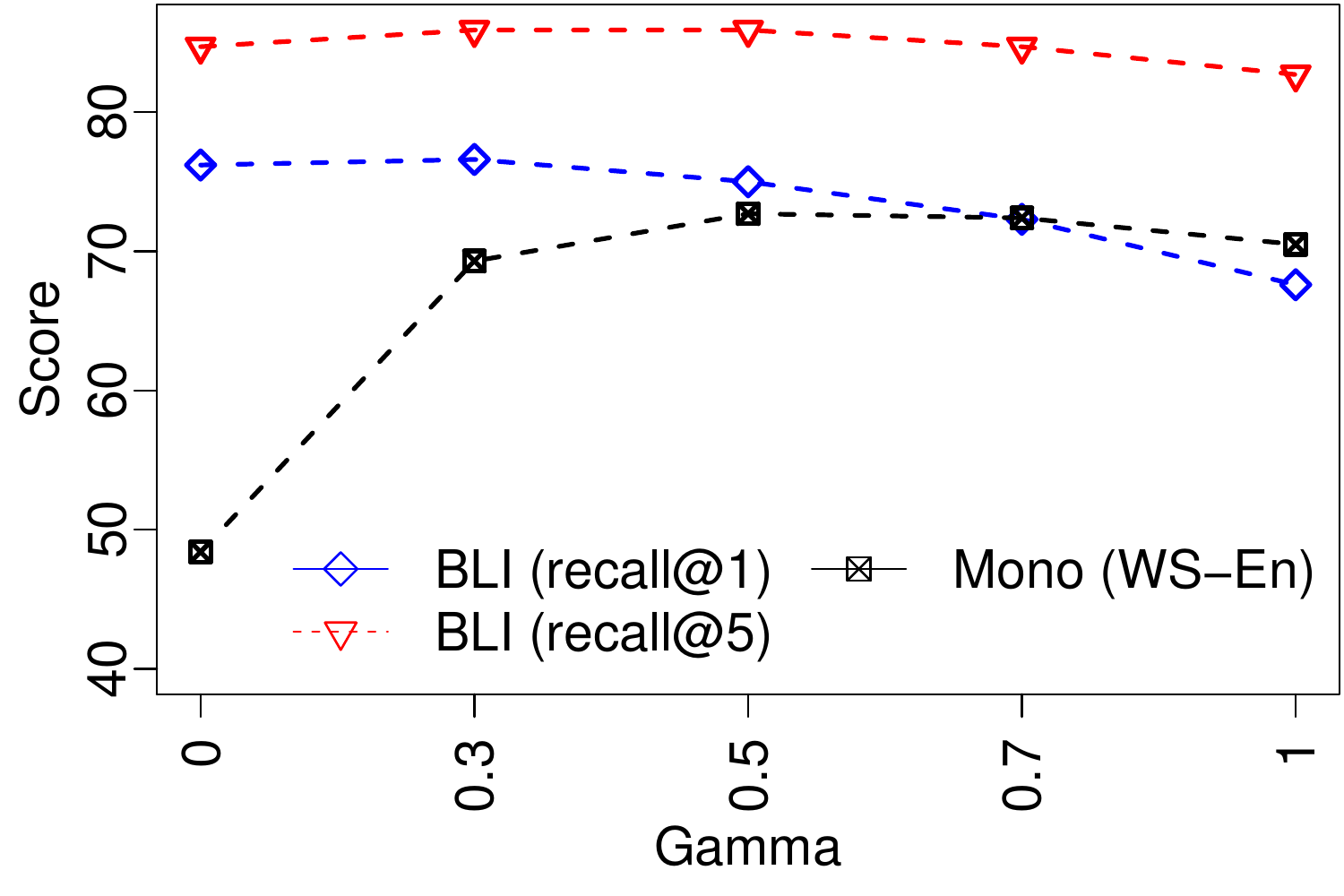}
 \caption{Performance of word embeddings interpolated using different values of $\gamma$ evaluated using BLI (Recall@1, Recall@5) and English monolingual WordSim353 (WS-En). }
 \label{fig:tune_gamma}
 \end{figure}

We use English - Italian pair for tuning purposes, considering the value of $\gamma$ in equation~\ref{equ:interpolate_combine}. 
Figure~\ref{fig:tune_gamma} shows the performances using different values of $\gamma$. 
The two extremes where $\gamma =0 $ and $\gamma = 1$ corresponds to no interpolation where we just use $\Mu$ or $\Mv$ respectively. 
As $\gamma$ increases, the performance on WS-En increases yet BLI  decreases. These results confirm our hypothesis in \S\ref{sec:combine_emb} that $\Mu$ is better at capturing bilingual relation and $\Mv$ is better at capturing monolingual relation. As a compromise, we choose $\gamma = 0.5$ in our experiments. Similarly, we tune the regularization sensitivity $\delta$ in equation~(\ref{equ:reg_combine}) which combines embeddings space during training. We test $\delta = 10^{-n}$ with $n = \{0,1,2,3,4\}$ and using $\Mv$, $\Mu$ or the interpolation of both $\frac{\Mv + \Mu}{2}$ as the learned embeddings, evaluated on the same BLI and WS-En. We select $\delta = 0.01$. 

\begin{table}
\centering
\resizebox{\columnwidth}{!}{
\begin{tabular}{llccc}
\toprule
&  \multirow{2}{*}{Model} & \multicolumn{2}{c}{BLI} & Mono  \\
&                               & $rec_1$ & $rec_5$ & WS-en \\
\midrule
\multirow{2}{*}{\rotatebox{90}{Alone}}& Joint-model + $\Mv$  & 67.6 & 82.8 & 70.5\\ 
& Joint-model + $\Mu$  & 76.2 & 84.7 & 48.4\\
\midrule  
\multirow{4}{*}{\rotatebox{90}{Combine}} & Interpolation   & 75.0 & 85.9 & 72.7  \\
& Concatenation                 & 72.7 & 85.2 & 71.2  \\
& Regularization + $\Mv $ & 78.8 & 88.6 & 51.1 \\
& Regularization + $\Mu $ & 78.7 & 89.5 & \textbf{75.4} \\
& Regularization + $\frac{\Mv + \Mu}{2}$ & \textbf{78.9} & \textbf{90.5} & 73.0 \\
\bottomrule
\end{tabular}
}
\caption{Performance on English-Italian BLI and English monolingual similarity WordSim353 (WS-en) for various combining algorithms mentioned in \S\ref{sec:combine_emb} w.r.t just using $\Mu$ or $\Mv$ alone (after joint-training). We use $\gamma = 0.5$ for interpolation and $\delta = 0.01$ for regularization with the choice of $\Mv$, $\Mu$ or combination of both$\frac{\Mv + \Mu}{2}$ for the output. The best scores are bold.}
\label{tab:performance_combine}
\end{table}

Table~\ref{tab:performance_combine} shows the performance with and without using combining algorithms mentioned in \S\ref{sec:combine_emb}. 
As the compromise between both monolingual and crosslingual tasks,  we choose regularization + $\Mu$ as the combination algorithm. 
All in all, we apply the regularization algorithm for combining $\Mv$ and $\Mu$ with $\delta = 0.01$ and $\Mu$ as the output for all language pairs without further tuning. 
 
\section{Crosslingual Document Classification}
In this section, we evaluate our CLWE on a downstream crosslingual document classification (CLDC) task.
In this task, the document classifier is trained on a source language and then applied directly to classify a document in the target language. 
This is convenient for a target low-resource language where we do not have document annotations. 
The experimental setup is from~\newcite{klementiev-titov-bhattarai:2012}.\footnote{The data split and code are kindly provided by the authors.}
The train and test data are from Reuter RCV1/RCV2 corpus~\cite{Lewis:2004:RNB}. 

The documents are represented as the bag of word embeddings weighted by \texttt{tf.idf}. A multi-class classifier is trained using the average perceptron algorithm on 1000 documents in the source language and tested on 5000 documents in the target language. We use the CLWE, such that the document representation in the target language embeddings is in the same space with the source language. 

\begin{table}
\centering
\tabcolsep 2pt
\resizebox{\columnwidth}{!}{
\begin{tabular}{lccc}
\toprule
Model         & $en\rightarrow de$ & $de \rightarrow en$ & Avg. \\
\midrule
MT baseline   & 68.1    & 67.4    & 67.8 \\
\newcite{klementiev-titov-bhattarai:2012} & 77.6    & 71.1    & 74.4 \\
\newcite{icml2015_gouws15}       & 86.5    & 75.0    & 80.8 \\
\newcite{kovcisky-hermann-blunsom:2014:P14-2}          & 83.1    & 75.4    & 79.3 \\
\newcite{Chandar-nips-14}        & \textbf{91.8}    & 74.2    & 83.0 \\
\newcite{DBLP:journals/corr/HermannB14}       & 86.4    & 74.7    & 80.6 \\
\newcite{Luong-etal:naacl15:bivec}        & 88.4    & \textbf{80.3}    & \textbf{84.4} \\
Our model     & 87.8    & 75.1    & 81.5 \\
\bottomrule
\end{tabular}
}
\caption{CLDC performance for both $en\rightarrow de$ and $de \rightarrow en$ direction for many CLWE. MT baseline uses phrase-based statistical machine translation to translate the source language to target language. 
The best scores are bold.}
\label{tab:cldc}
\end{table}    

We build the en-de CLWE using combined models as described in equation~(\ref{equ:reg_combine}). Following prior work, we also use monolingual data from the RCV1/RCV2 corpus~\cite{klementiev-titov-bhattarai:2012,icml2015_gouws15,Chandar-nips-14}.\footnote{We randomly sample documents in RCV1 and RCV2 corpus and selected around 85k documents to form 400k monolingual sentences for both English and German.
For each document, we perform basic processing including: lower-case, remove tags and tokenize. These monolingual data are then concatenated with the monolingual data from Wikipedia to form the final training data.}

Table~\ref{tab:cldc} shows the CLDC results for various CLWE. Despite its simplicity, our model achieves competitive performance. Note that aside from our model, all other models in Table~\ref{tab:cldc} use a large bitext (Europarl) which may not exist for many low-resource languages, limiting their applicability. 

\section{Conclusion}
Previous CLWE methods often impose high resource requirements yet have low accuracy. We introduce a simple framework based on a large noisy dictionary. We model polysemy using EM translation selection during training to learn bilingual correspondences from monolingual corpora. Our algorithm allows to train on massive amount of monolingual data efficiently, representing monolingual and bilingual properties of language. This allows us to achieve state-of-the-art performance on bilingual lexicon induction task, competitive result on monolingual word similarity and crosslingual document classification task. Our combination techniques during training, especially using regularization, are highly effective and could be used to improve monolingual word embeddings. 

\bibliography{References}
\bibliographystyle{emnlp2016}

\end{document}